\documentclass[letterpaper]{article}
\usepackage{aaai2027}
\usepackage[hyphens]{url}
\usepackage{graphicx}
\usepackage{natbib}
\usepackage{caption}
\usepackage{amsmath}
\usepackage{booktabs}
\usepackage{tabularx}
\usepackage{fvextra}
\DefineVerbatimEnvironment{PromptBlock}{Verbatim}{
  fontsize=\scriptsize,
  breaklines=true,
  breakanywhere=true,
  breaksymbolleft={}
}

\usepackage{algorithm}
\usepackage{algorithmic}
\usepackage{xcolor}
\usepackage{colortbl}
\definecolor{ourshade}{gray}{0.93}
\title{DiagLoop: A Counterfactual Data Flywheel with Stage-Localized\\Reinforcement for Diagnostic LLMs}
\author{Jian Zhang, Bingyi Wang, Yizhi Liu}
\affiliations{Zhejiang University}
\nocopyright
\begin{document}

\maketitle

\begin{abstract}
Causal diagnostic models must justify how their conclusions follow from the evidence because diagnoses guide repairs and treatments. Sensitive data and latency constraints also favor local deployment. Training a locally deployable LLM is difficult, however: serious cases are scarce, records rarely contain reasoning paths, and data from one configuration transfer poorly to another. We present DiagLoop, which turns codified physical relations or clinical guidelines, authored once per mechanism family, into supervision beyond recorded cases. A training-only teacher proposes counterfactual worlds by varying causes, contexts, and observations, while a separate hybrid checker decides which worlds are admissible. A student model then answers each world through symptom abstraction, causal-chain construction, and root-cause attribution. Criteria attached to these stages locate the student's earliest failure; for a nonterminal failure, a bounded repair probes the model's downstream competence, and the resulting weakness profile routes subsequent generation. Stage-localized reinforcement learning rewards only the model-generated continuation, while replay and preservation limit forgetting. The same stage criteria govern admission, attribution, reward, and regeneration through operationally separate checks that the proposer never performs. Trained only on synthesized scenarios and without case-level expert reasoning annotations, the resulting 8B model improves strict path correctness by 11.6 and 5.5 points over the strongest conventional baseline across eight industrial systems and ten disease categories, and by 3.9 and 2.3 points over a deranged-routing control. It also exceeds the evaluated proprietary references in path correctness in both domains, even when those references receive few-shot examples or the specification in context.
\end{abstract}

\section{Introduction}


Industrial fault diagnosis and clinical diagnosis share one core task: inferring the hidden cause behind the observed evidence, and LLMs are increasingly applied to both \citep{appliedenergy2024fault,lin2025fdllm,aei2026fault,singhal2023large,mcduff2025towards}. A diagnosis is acted on: equipment is repaired and treatment is started on the strength of the reasoning behind the conclusion, so the reasoning path must be right, not only the final label. Diagnostic data are also sensitive and often processed under network-isolation and latency constraints \citep{dennstadt2025implementing}; practice therefore needs a readily deployable open LLM whose conclusions are supported by checkable reasoning.

Training such a model runs into the data first. Serious faults are rare, can disrupt the very streams that record them, and cannot be induced safely; clinical records are constrained by privacy and limited sharing. Each archive is also bound to one installation or cohort, so models trained on it learn setting-specific patterns rather than knowledge that transfers \citep{liu2021transfer}. Most records, moreover, keep only the diagnostic conclusion, not the reasoning that established it.

An experienced engineer or physician, however, can often diagnose a device or a patient never seen before. Expert reasoning runs on the relations between manifestations and faults or diseases, and these relations are far more stable across installations and cohorts than the surface patterns of any single archive. Given a candidate cause, an expert asks which manifestations should follow and checks them against the case; look-alike causes are told apart by what should differ, and changed conditions by how the observations should shift. This is counterfactual reasoning in the practical sense: comparing the case at hand against controlled alternative worlds, the same fault under a different condition, a different fault behind similar evidence. This capability is exactly what we want the model to acquire: the mechanism-grounded path from evidence, through a causal chain, to a cause, and the ability to tell such alternative worlds apart. The supervision it requires follows directly: families of controlled scenario variants that differ in one cause or one condition at a time, with reasoning paths checkable stage by stage---symptom abstraction, causal-chain construction, and root-cause attribution.

Supervision of this form is hardly found in records and has to be synthesized. Earlier augmentation, such as GAN-based generation, enlarges the dataset around the recorded distribution \citep{shao2019generative,ma2021interpretable} but rarely creates scenarios beyond it, let alone the reasoning behind them. Recent work therefore uses LLMs to write the training data itself, from instructions and rationales to progressively harder tasks \citep{wang2023selfinstruct,zelikman2022star,xu2024wizardlm,gunasekar2023textbooks}. The trouble is that the generating model itself makes causal errors: a generated scenario or its reasoning chain may contradict the physical or clinical relations, omit the evidence needed for diagnosis, or leave a look-alike cause equally supported. The usual filters---answer agreement, consensus, outcome correctness, or task-level rubrics \citep{lyu2026agenticqwen,lyu2026synthagent}---are themselves model judgments prone to the same causal errors, so flawed worlds pass. What training needs, then, is synthetic supervision whose scenarios and reasoning are checked for consistency with the governing mechanism itself, a check that existing pipelines have yet to provide, the \emph{verification gap}.

At the same time, a trustworthy corpus alone does not finish the job, because training keeps changing the model. Mastered scenarios provide diminishing signal; rollout-based training draws its signal precisely from what the model still gets wrong \citep{shao2024deepseekmath,lyu2026synthagent}. Further progress depends on finding where the current model fails and aiming new data exactly there, and locating that place is the hard part: whether an answer is right or wrong is too coarse a signal, since in chain-shaped diagnosis one early mistake can invalidate every later stage, and the final result reveals neither the stage that broke nor the kind of variation that exposed it. Existing tools do not close this loop either: process supervision scores intermediate steps, but the scores end there \citep{lightman2024lets,wang2024mathshepherd}; curricula reorder examples that already exist \citep{bengio2009curriculum}; error-driven generation expands failed tasks without locating the failing stage \citep{lyu2026agenticqwen}. What is still missing is a path from an observed weakness to new data that targets it, the \emph{attribution gap}; and once targeted updates are made, shared parameters let progress on a weak stage erode stages that were already right, the \emph{consolidation gap} \citep{kirkpatrick2017overcoming}.

We therefore propose \textbf{DiagLoop}, which trains a locally deployable diagnostic LLM by coupling counterfactual data construction, causal weakness mining, and stage-localized reinforcement learning in one closed loop. Its central design is criterion reuse: a training-only teacher proposes worlds and bounded repairs but never certifies its own output, while the same three-stage criterion semantics, implemented through operationally separate checks, govern admission, failure attribution, reward, and regeneration. At inference, only the model and output schema remain. Our contributions are:
\begin{itemize}
    \item \textbf{Counterfactual world flywheel.} It turns codified domain specifications into diverse counterfactual scenarios, and admits a scenario only when its reasoning path passes checks against the same codified mechanism, so training needs no case-level expert reasoning annotations.
    \item \textbf{Causal weakness flywheel.} The same stage criteria locate the earliest observable failure in each model trajectory; a checker-admitted bounded repair followed by the model's own continuation shows whether the failure stays local or propagates, and the resulting weakness profile decides which scenarios are generated next.
    \item \textbf{Stage-localized RL.} The located failure then sets where the rollout starts and which criteria are rewarded on the model's own continuation, while replay, preservation trajectories, and a reference anchor keep mastered skills intact. Trained solely on synthesized scenarios, the resulting 8B model surpasses the strongest conventional baseline, the deranged-routing control, and the evaluated zero-shot proprietary references in strict path correctness across eight industrial systems and ten disease categories.
\end{itemize}

\section{Related Work}


\subsection{LLMs for Industrial and Clinical Diagnosis}

Applied LLM diagnosis increasingly targets locally deployable open models, since evidence often cannot leave isolated networks, monitoring bounds latency, and conclusions should be auditable on site \citep{lin2025fdllm,appliedenergy2024fault,aei2026fault,singhal2023large,mcduff2025towards,dennstadt2025implementing}. The prevailing recipe fine-tunes on fault or disease labels, at most using diagnosis accuracy to trigger further data generation \citep{appliedenergy2024fault,aei2026fault}. The cost of this label focus is documented: a model can return the correct fault label with an incorrect explanation, and improving diagnosis accuracy does not ensure inference correctness \citep{appliedenergy2024fault}. Augmentation and transfer learning reduce label scarcity \citep{shao2019generative,ma2021interpretable,liu2021transfer}, yet labels neither block setting-specific shortcuts \citep{geirhos2020shortcut} nor recover unrecorded reasoning paths. What this line lacks is supervision over how the model reasons; the codified rules and guidelines these domains already maintain \citep{katipamula2005methods,sutton2020overview} are a natural standard for such supervision, yet current LLM diagnosis training leaves them unused.

\subsection{Counterfactual and Verifiable Synthetic Data}

Diagnosis needs specific synthetic supervision: scenarios that vary causes and conditions, each with a reasoning path trustworthy enough to train on. Existing synthesis scales volume and difficulty, creating instructions, rationales, tasks, and corpora for smaller models \citep{wang2023selfinstruct,zelikman2022star,xu2024wizardlm,gunasekar2023textbooks}, expanding failed tool-use tasks under answer agreement \citep{lyu2026agenticqwen}, and scoring generated ecosystems with task-level rubrics \citep{lyu2026synthagent}. None of this makes the reasoning reliable: generators commit causal errors, and agreement- or rubric-based filters---model judgments often from related families---pass many errors on, so the paths that reach training cannot be presumed correct. Rule-derived benchmarks and causal probes carry more reliable structure \citep{tchango2022ddxplus,frohberg2022crass,kiciman2023causal} yet serve evaluation rather than training, and counterfactually edited examples improve robustness in training \citep{kaushik2020learning} but rely on human editing of recorded cases rather than mechanism-checked synthesis. A synthesis route that delivers mechanism-checked, trustworthy reasoning paths at training scale is still missing, the \emph{verification gap}.

\subsection{Process Supervision and Adaptive Updating}

A further line of work concerns where a model fails and how training should respond. Outcome rewards give a reliable end-of-chain signal \citep{uesato2022solving}, process reward models score intermediate steps \citep{lightman2024lets,wang2024mathshepherd,setlur2025rewarding}, curricula order examples from easy to hard \citep{bengio2009curriculum}, and error-driven flywheels grow the corpus around failed tasks \citep{lyu2026agenticqwen}. These signals are designed for generic steps and whole tasks; connecting a located failure to the next round of generation lies outside their scope, the \emph{attribution gap}. On the training side, critique-and-revision uses self-feedback \citep{madaan2023selfrefine}, on-policy distillation follows the model's own errors \citep{hinton2015distilling,agarwal2024onpolicy}, verifiable-reward RL optimizes checkable tasks \citep{shao2024deepseekmath,guo2025deepseekr1}, and replay or weight regularization stabilizes updates \citep{rolnick2019experience,kirkpatrick2017overcoming}. These methods optimize the trajectory as a whole; for a chain whose stages share one network, conditioning updates on the located failure while protecting the rest is not addressed jointly by these methods, the \emph{consolidation gap}.

\paragraph{Summary.}
Prior work advances synthetic data, process feedback, adaptive generation, and verifiable optimization separately, connecting them through different signals. The central contribution of \textbf{DiagLoop} is a \emph{criterion-reuse interface}: shared stage-criterion semantics govern admission, locate the earliest failure, score the model-generated continuation, route subsequent generation, and select passing trajectories for preservation, while the corresponding checks remain operationally separate.

\section{Formalization and Working Hypotheses}
\label{sec:hypothesis}

Within a codified mechanism family, how a condition manifests is governed far more by the codified relations than by the surface patterns of any particular installation or patient, and practice has already written these relations down. This yields our first hypothesis: \textbf{(H1)} the codified relations between manifestations and conditions are formalizable and compositional, so holding the relations fixed while varying conditions, contexts, and observations yields candidate counterfactual worlds whose validity the same codified relations can check. Concretely, let $\mathcal{F}$ denote the codified conditions, $\mathcal{C}$ the contexts (system forms, sensor suites, operating regimes, or patient profiles), and $\mathcal{V}$ a set of bounded variation operators on worlds and their observations. Composing at most $K$ operators over a base world yields the counterfactual space
\begin{equation}
\Omega_K=\{v_k\!\circ\!\cdots\!\circ\!v_1(\omega_0(f,c)):f\!\in\!\mathcal{F},\,c\!\in\!\mathcal{C},\,v_i\!\in\!\mathcal{V},\,k\!\le\!K\},
\label{eq:space}
\end{equation}
where $\omega_0(f,c)$ is a base world, composition operators intervene on the condition or context, and the remaining operator families perturb the observations. The number of candidate sequences grows exponentially with $K$; deduplication and admission set the realized scale (Section~\ref{sec:experiments}), while any single archive records a sparse slice.

Over this space the two directions of answering differ sharply in difficulty. Asking an LLM to diagnose directly is hard: it must weigh every candidate in $\mathcal{F}$ against overlapping evidence and discharge the differentials among look-alikes. Constructing with the answer in hand is far more constrained, though not error-free: given the condition, laying out its manifestations and the connecting reasoning path follows the codified relations $\mathcal{R}_f$, a substantially narrower construction problem even when multiple valid derivations exist. This asymmetry mirrors expert practice and is an empirical matter, stated as our second hypothesis: \textbf{(H2)} for the same LLM, constructing a scenario and its reasoning path with the condition known is substantially more reliable than inferring the condition by direct diagnosis. Section~\ref{sec:experiments} tests both directions for the same teacher.

\section{Method}
\label{sec:method}

\begin{figure*}[t]
\centering
\includegraphics[width=0.94\textwidth,height=0.3774\textwidth,trim=0 3.5 0 3.5,clip]{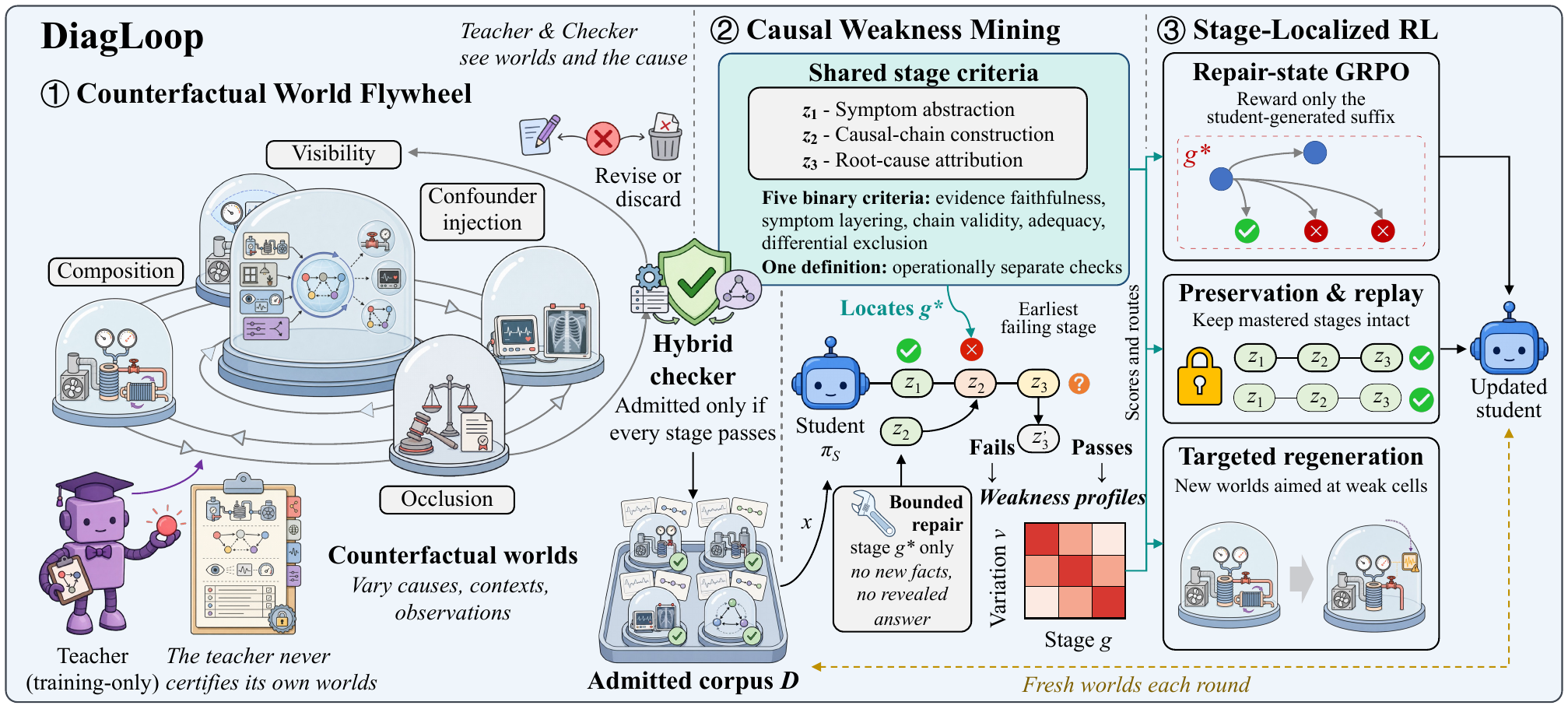}
\caption{DiagLoop overview: one set of stage criteria connects admission, attribution, reward, and regeneration.}
\label{fig:overview}
\end{figure*}

DiagLoop turns the knowledge that a domain already records into training for a locally deployable diagnostic student: admitted counterfactual worlds supply the data, the earliest failure supplies the target, and stage-localized reinforcement learning supplies the update. The teacher and checker see this knowledge and the intended cause; the student sees only observations $x$, and only the student and output schema remain at inference. One set of stage criteria, defined next, governs admission, attribution, reward, and regeneration through operationally separate checks (Figure~\ref{fig:overview}); generalization claims are limited to new configurations and compositions within the codified family.

\subsection{World Specification and Stage Criteria}
\label{sec:specification}

A world specification $\mathcal{W}$ records the system structure (components, topology, and observable variables), its operating logic, and the codified relations $\mathcal{R}_f$ from each condition to its manifestations. Logic and mechanisms are authored once per domain; structural fields instantiate each setting. An admitted world $\omega=(f,c,x_\omega)$ binds a condition, a context, and its teacher-proposed observations to the specification. The specification encodes physical constraints in the industrial domain and guideline relations in the clinical domain; the method requires a codified mechanism family and an observation-to-field mapping (authoring costs in Appendix~A).

Given observations $x$, the student produces $\tau=(z_1,z_2,z_3)$: symptom abstraction forms an evidence set, causal-chain construction links it to candidate conditions, and root-cause attribution selects a cause while excluding alternatives. Five binary criteria are assigned by stage: $\mathcal{K}_1=\{$evidence faithfulness, symptom layering$\}$, $\mathcal{K}_2=\{$causal-chain validity$\}$, and $\mathcal{K}_3=\{$root-cause adequacy, differential exclusion$\}$. Stage $g$ passes when $P_g(\tau;\omega,\mathcal{W})=\prod_{\kappa\in\mathcal{K}_g}\kappa(z_{1:g};\omega,\mathcal{W})=1$; each criterion is assigned to one stage but may read the preceding stages. For a cooling-coil valve stuck closed, for example, the checks reject an invented flow reading or a conclusion that retains a fouled coil even though fouling cannot explain valve-position feedback that still reads closed despite an opening command.

The checker is hybrid: specification lookup resolves structured questions, and an LLM only maps free-form spans to fields in a context separate from the teacher's proposal pass. Each violation cites its field and supporting span. This bounded checker is not a formal verifier, so consistency claims are conditioned on expert-audited false-acceptance and false-rejection rates. When scoring a student trajectory, it receives the intended cause and alternatives, so it maps, compares, and looks up rather than diagnoses. Test evaluation uses independent gold chains and a frozen scorer; the training checker never scores test outputs.

\subsection{Counterfactual World Flywheel}
\label{sec:world-flywheel}

Following H2 (Section~\ref{sec:hypothesis}), the world-construction loop fixes the cause, and hence the intended label, first. DiagLoop therefore never asks the teacher to diagnose within this loop; it asks the teacher to construct the cause's consequences. The student sees only $x_\omega$ and still solves the full inverse problem.

The four operators in $\mathcal{V}$ instantiate the space $\Omega_K$ of Eq.~\eqref{eq:space}. Composition creates a new anchor by placing $f$ in a rare context or replacing it with intervention-defined $f'$; the label-changing subset, its target fixed before generation, feeds the intervention-sensitivity evaluation (Section~\ref{sec:experiments}). Visibility, confounder injection, and occlusion preserve the underlying condition and diagnosis while changing which readings are exposed, added, or hidden, with controlled pairs otherwise fixed. We use \emph{counterfactual} for these world interventions and observation perturbations, without claiming unit-level causal counterfactuals \citep{pearl2009causality}. Later allocation under budget $K$ follows the weakness profile, not $k$ alone.

For world $\omega$, teacher policy $\pi_T$ proposes observations $x_\omega$ and reference trajectory $\tau_\omega$. The admitted corpus is
\begin{equation}
\mathcal{D}=\{(x_\omega,\tau_\omega)\sim\pi_T(\cdot\mid\omega,\mathcal{W}):\omega\in\Omega_K,\ \tau_\omega\models\mathcal{K}\},
\label{eq:corpus}
\end{equation}
where $\tau_\omega\models\mathcal{K}$ means that every stage passes. Failed drafts are minimally revised and rechecked, and irreparable worlds are discarded. Admission jointly requires a diagnosable world, evidence sufficient for a valid derivation, a fully passing reference path, and explicit exclusion of relevant look-alikes. Clinical checks enforce guideline-supported admissibility rather than physical necessity. The training pair is $(x_\omega,\tau_\omega)$; audit records additionally store the intended cause, variation metadata, and criterion results.

\subsection{Causal Weakness Flywheel}
\label{sec:weakness}

The student first answers without assistance. The checker then identifies its earliest failing stage,
\begin{equation}
g^{*}(\tau)=\min\bigl(\{\,g : P_g(\tau;\omega,\mathcal{W})=0\,\}\cup\{\infty\}\bigr).
\label{eq:gstar}
\end{equation}
Attribution stops at the earliest observable failure, since an earlier violation invalidates the premises of every later stage. If all stages pass, $g^{*}=\infty$ and no repair is made. This localization is criterion-relative, not a claim about the parameter-level origin of the error.

For $g^{*}<3$, chain repair rewrites only the failing stage and is admitted only if that stage passes; it may not add unobserved facts, reveal the intended cause, or provide a complete answer, with field--span checks screening leakage. The retained student prefix is $z_{1:g^{*}-1}$ (empty when $g^{*}=1$), the admitted repair replaces $z_{g^{*}}$, and the student generates $z'_{g^{*}+1:3}$. The repair enters the prompt but never the loss. A passing continuation is consistent with a stage-local bottleneck; a failed continuation enters the post-repair failure profile. Further repairs are profiling probes only and never enter the reinforcement-learning rollout, so each episode has one failure boundary. At terminal $g^{*}=3$, no repair is supplied and the student regenerates $z'_3$ from $z_{1:2}$. We aggregate first-failure and post-repair failure rates by stage $g$ and variation family $v$: $w_t(g,v)=\alpha\,\mathrm{FF}_t(g,v)+(1-\alpha)\,\mathrm{PRF}_t(g,v)$, with $\alpha\in[0,1]$. This profile allocates the next generation round.

\subsection{Stage-Localized Reinforcement Learning}
\label{sec:localized-rl}

Training starts with one SFT pass over $\mathcal{D}$, then alternates failure-boundary-conditioned reinforcement learning, preservation, and regeneration; \emph{stage-localized} names the rollout boundary, not stage-specific parameters. For $g^{*}<3$, recovery at the failing stage is not assumed and is measured on later unassisted answers.

\textbf{Reinforcement learning from the repaired boundary.} Multiple continuations may be valid, so we optimize the criterion-scored suffix with GRPO \citep{shao2024deepseekmath}. For $g^{*}<3$, the prompt contains the student's stages before the failure and the admitted replacement for $z_{g^{*}}$; the student generates only the later stages. For $g^{*}=3$, it receives no repair and regenerates stage 3. With $b(g^{*})=\min\{g^{*}+1,3\}$, the reward is
\begin{equation}
R(\tau')=\sum_{g=b(g^{*})}^{3}\ \sum_{\kappa\in\mathcal{K}_g}\lambda_\kappa\,\kappa(z'_g;\omega,\mathcal{W})\;-\;\mu\,U(\tau').
\label{eq:reward}
\end{equation}
Only student-generated stages receive credit: repaired stages are excluded and terminal stage 3 is included; both $g^{*}\!=\!2$ and $g^{*}\!=\!3$ reward only stage 3, differing in whether a repair conditions the prompt. $\lambda_\kappa$ weights the criteria and is frozen before training, largest on causal-chain validity and root-cause adequacy; the vector and an equal-weight sensitivity appear in the appendix. $U(\tau')$ penalizes assertions unsupported by $x$ or $\mathcal{W}$, scored by the same checking pass as the criteria. GRPO normalizes advantages within groups against the post-SFT KL reference; rounds stop when the development gain falls below a frozen threshold, capped at eight rounds, and both domains stopped after six. The primary stream is on-policy: fresh unassisted rollouts from $\pi_{\theta_t}$ and their repair-conditioned continuations rebuild the round-$t$ failure pool and $w_t$; earlier rollouts enter only through replay.

\textbf{Preservation.} Each batch also includes full trajectories from cells below a frozen weakness threshold, replay from earlier rounds, and a small SFT term on checker-admitted teacher references, limiting cross-stage interference, drift, and format collapse. Full preservation trajectories ($g^{*}=\infty$) score all three student-generated stages directly.

\textbf{Regeneration.} Before training, we freeze an executable map $\rho(g,v)$ for every stage--variation cell, specifying the target criterion or mixture, operator and subtype, invariant fields, and admission check. After each update,
\begin{equation}
q_{t+1}(g,v)=(1-\epsilon)\,\frac{\exp(\eta\,w_t(g,v))}{\sum_{g',v'}\exp(\eta\,w_t(g',v'))}+\epsilon\,q_0(g,v),
\label{eq:regen}
\end{equation}
where $\eta$ controls concentration, $\epsilon\in[0,1]$, and uniform $q_0$ preserves support for every cell. The generator samples a cell, applies $\rho(g,v)$, and rechecks the variant, creating new supervision rather than resampling a static pool \citep{lyu2026synthagent}. Admitted reference trajectories from these newly routed worlds enter the next round's SFT-anchor stream: this loss covers the full path and is targeted only through data routing, and it supplies the located weak stage the direct supervision that repair-conditioned rollouts withhold.

Admission, failure localization, and reward scoring use separate calls with role-specific prompts under shared criterion definitions; test scoring uses a frozen, independently constructed scorer. Appendix~A provides specifications, operators, $\rho$, prompts, model versions, hyperparameters, the stopping rule, and seeds.

\section{Experiments}
\label{sec:experiments}

\subsection{Experimental Setup}
\label{sec:setup}

\paragraph{Benchmarks and data separation.}
The LBNL industrial fault benchmark \citep{granderson2022lbnl} contains 26{,}175 cases covering 91 fault types in eight heterogeneous systems; our DDXPlus evaluation subset \citep{tchango2022ddxplus} contains 128{,}800 cases grouped into ten disease categories. Results are macro-averaged at the group level, with per-group breakdowns in Appendix~D; paired values read industrial/clinical throughout. Neither benchmark contributes cases to synthesis or training: world specifications derive from separately frozen standards or guidelines, both evaluations hold out cases, and LBNL adds a configuration-held-out split fixed before synthesis. A disjoint synthetic development set fixes prompts, thresholds, hyperparameters, stopping targets, and checkpoints before either test set or the locked audit is accessed. Selection rules, distributions, provenance, and ontology overlap appear in Appendix~B; specification versions and authoring effort appear in Appendix~A.

\paragraph{Models.}
The student is Qwen3-8B, trained and served on the same 4$\times$NVIDIA H200 cluster; a single card suffices for deployment at 3.6~s per case (bf16, 2{,}048-token cap). Frozen Qwen3.7-Max \citep{qwen37} proposes and repairs training scenarios; at evaluation it serves only as a zero-shot reference without the intended cause or specification; its undisclosed parameter count precludes model-size claims. Claude-Sonnet-4.6 is a second reference (API details in Appendix~B). Algorithm, stage schema, and optimization hyperparameters are shared across domains; only the specification, field mapping, and criterion instantiation change. Counterfactual training worlds are enumerated forward from the frozen specifications: four perturbation operators (occlusion, confounder, visibility, composition) at depth $K{=}3$ over 99 industrial conditions (91 faults plus eight normals; ${\sim}$30 sensors, nine operating contexts) and 49 clinical conditions (223 symptoms, demographic grid) yield 176{,}891/131{,}742 deduplicated combinations, of which physical and clinical filtering retains 30{,}614/49{,}876 generable worlds (operator combinatorics and per-round yields in Appendix~B). Training runs six flywheel rounds per domain over these worlds, admitting 14{,}302/21{,}744 scenarios at 218/342 aggregate GPU-hours, of which 12{,}874/19{,}566 (excluding development and profiling reserves) supply cold-start SFT answers.

\paragraph{Evaluation and statistics.}
Gold reasoning chains are independent of training: clinical chains derive from benchmark evidence and differential annotations, and domain professionals annotate every industrial evaluation case. A sealed 1{,}500-scenario audit, stratified across admitted and rejected drafts and excluded from prompt tuning, estimates the training checker's error and $g^{*}$-localization rates. Test paths are scored by a frozen gold-chain-conditioned scorer whose agreement is measured against method-blinded experts; the training checker never grades test output, and expert $g^{*}$ labels agree at Cohen's $\kappa=0.81$ on a 300-case double-annotated subset (Appendices B and C).

We report top-1 accuracy (Acc), strict all-stage path correctness (Path), and stage-level scores, each averaged over five sampled completions per case at temperature 0.6 for trained models; macro-F1 appears in Appendix~D. Main comparisons average three seeds, the two prespecified factorials use five paired seeds each, paired bootstrap preserves method pairing, and test families with Holm correction were fixed before evaluation (resampling details in Appendix~B).

\paragraph{Baselines and budgets.}
At evaluation, all models share evidence, candidate space, schema, parser, and tool access. Data baselines fine-tune Qwen3-8B on unfiltered, answer-agreement-filtered, or constraint-admitted teacher scenarios; plain GRPO on uniformly sampled fresh worlds is the compute-matched optimization baseline. The routing ablation adds a uniform-regeneration arm: the proposal distribution is frozen to uniform while fresh worlds are still generated, with calls, tokens, and GPU-hours matched (Appendix~D). A type-compatible derangement of complete $\rho$ entries forms the deranged-routing control, which decouples measured weaknesses from their assigned generation operations (staleness robustness in Appendix~G). Main-table comparisons match generated and admitted scenarios, student tokens and updates, teacher and checker calls, and replay settings; the admission ablation instead equalizes admitted counts to isolate quality from size.

\subsection{Teacher Premise and Main Results}
\label{sec:premise}

\paragraph{Forward construction.}
Testing H2, Qwen3.7-Max performs both directions on the matched development sets: direct diagnosis passes the stage criteria in 41.6\%/38.2\% of cases. Given the intended cause, first drafts pass in 78.4\%/74.9\% and revise--recheck reaches admission within 3 rounds in 93.1\%/90.6\%; expert audit preserves this ordering, supporting H2 (revision and call costs in Appendix~C). Admission rates stay within 5.1 points of the depth-one rate through $k\!=\!K$, an operational feasibility check for H1 (Appendix~C); the operator and head--tail results below assess the utility of the resulting variations.

\begin{table}[t]
\centering
\footnotesize
\renewcommand{\arraystretch}{1.0}%
\setlength{\tabcolsep}{2.5pt}
\begin{tabularx}{\columnwidth}{@{}Xcccc@{}}
\toprule
& \multicolumn{2}{c}{\textbf{LBNL (8 sys.)}} & \multicolumn{2}{c}{\textbf{DDXPlus (10 cat.)}} \\
\cmidrule(lr){2-3}\cmidrule(lr){4-5}
Method & Acc & Path & Acc & Path \\
\midrule
\multicolumn{5}{c}{\textbf{Proprietary references (zero-shot)}} \\
\midrule
Qwen3.7-Max & 77.80 & 70.47 & 78.74 & 54.08 \\
Claude-Sonnet-4.6 & 86.05 & 72.60 & 82.34 & 57.08 \\
\midrule
\multicolumn{5}{c}{\textbf{Student: Qwen3-8B}} \\
\midrule
Qwen3-8B (zero-shot) & 46.86 & 14.31 & 51.24 & 16.48 \\
SFT (teacher, unfiltered) & 83.94 & 70.07 & 75.14 & 60.59 \\
SFT (+answer agreement) & 86.55 & 74.50 & 77.24 & 61.78 \\
SFT (constraint-admitted) & 90.26 & 80.37 & 78.85 & 64.78 \\
Plain GRPO (uniform fresh) & 89.53 & 79.05 & 78.44 & 64.08 \\
DiagLoop (deranged routing) & 92.45 & 88.07 & 78.37 & 67.98 \\
\rowcolor{ourshade}
\textbf{DiagLoop (full)} & \textbf{94.66} & \textbf{91.97} & \textbf{79.84} & \textbf{70.23} \\
\bottomrule
\end{tabularx}
\caption{Main results (\%): top-1 accuracy (Acc) and strict all-stage path correctness (Path), macro-averaged over eight systems or ten disease categories; best trained method in bold. Trained methods use matched budgets (Section~5.1); per-group results appear in Appendices D--G.}
\label{tab:main}
\end{table}

\paragraph{End-to-end performance.}
Table~\ref{tab:main} yields three findings. First, DiagLoop reaches 94.66/91.97 Acc/Path on LBNL and 79.84/70.23 on DDXPlus: Path improves by 11.6 and 5.5 points over constraint-admitted SFT, the strongest conventional baseline (paired 95\% CIs [8.9, 13.6] and [4.7, 7.9]), and by 3.9/2.3 points over the deranged-routing control, the closest arm overall. Second, the gains are broad rather than driven by a few groups: improvements occur in 8/8 systems and 9/10 categories, and the worst-group changes are $+2.3$ (industrial) and $-1.4$ (clinical) percentage points, with the single clinical regression in the symptom/sign category (reversal analysis in Appendix~D). Third, on the configuration-held-out LBNL split, the Acc/Path gains over constraint-admitted SFT and plain GRPO are 3.4/8.2 and 3.9/8.8 points, so the closed loop adds capability beyond admission quality alone and beyond familiarity with seen configurations (every arm degrades on this split, ordered by configuration exposure; Appendix~D). These orderings are not a scorer-family artifact: a Claude-based scorer preserves the ranking exactly, per-arm 200-case blinded expert audits (separate from the sealed 1{,}500-scenario audit) agree with the frozen scorer on 91.5\%/88.5\% of cases, and full-arm scorer bias is statistically indistinguishable from zero (Appendix~E).

The proprietary APIs serve as capability references under the same output schema, not budget-matched comparisons; GPU memory and latency, rather than API accuracy, characterize deployability. On DDXPlus, Claude keeps a higher top-1 accuracy (82.34 vs.\ 79.84) yet trails by 13.2 Path points; its Acc--Path gap of 25.3 points, against 9.6 for DiagLoop, quantifies the right-label, wrong-path failure that motivates this work. The margin over the teacher reflects H2, not model size: the student distills the teacher's reliable fixed-cause constructions into inverse-task competence that the teacher itself lacks in the zero-shot setting; few-shot and specification-in-context references narrow but do not close the gap, remaining 11.3/7.8 Path points below under the pooled protocol (Appendix~E).

\subsection{Component Analyses}
\label{sec:mechanistic}

\paragraph{Constraint-based admission.}
Table~\ref{tab:ledger} prices each reuse of the stage criteria by its own control; the estimands are not additive. The deranged control retains every component and budget and breaks only the weakness-to-operation alignment, so its 3.9/2.3-point margin isolates the coupling itself, on top of the 11.6/5.5-point margin over the strongest conventional baseline. Every admission rule sees the same raw candidate pool; downstream sets are cut to the smallest admitted count and matched in tokens and updates, separating quality from size, with yield analyzed separately. Relative to answer agreement, constraint-based admission increases expert-audited validity by 7.8/6.4 points and downstream Path by 5.9/3.0 points. On the sealed 1{,}500-scenario audit, the checker falsely accepts 4.25/5.11\% and falsely rejects 6.83/7.98\% of scenarios, without amplification across training rounds (Appendix~C), and locates $g^{*}$ correctly in 87.35/84.63\% of cases; the frozen test scorer matches method-blinded experts on 91.17/89.50\%. Replacing the training checker with a disjoint model costs 1.4/1.2 admission-rate and 1.2/1.0 retrained Path points; replacing the proposer with the weaker Qwen3.6-Plus costs only 3.8/4.5 admission-rate and 1.4/1.5 Path points despite an 8.6/9.2-point first-draft drop: the loop absorbs quality differences on either side (Appendices C and E).

\begin{table}[t]
\centering
\footnotesize
\setlength{\tabcolsep}{3pt}
\begin{tabularx}{\columnwidth}{@{}l>{\raggedright\arraybackslash}Xcc@{}}
\toprule
Criterion reuse & Control & Ind. & Clin. \\
\midrule
Admission & vs.\ answer agreement & $+5.9$ & $+3.0$ \\
Attribution & repair-state episode & $+4.6$ & $+3.8$ \\
Reward & criterion scope & $+3.1$ & $+2.6$ \\
Regeneration & vs.\ deranged routing & $+3.9$ & $+2.3$ \\
Coupling & $2{\times}2$ interaction $\Delta_{\rm int}$ & $+2.4$ & $+1.9$ \\
\bottomrule
\end{tabularx}
\caption{Each reuse of the stage criteria, priced by its own control ($\Delta$Path, points); estimands are not additive. Admission and routing rows are macro-averaged (recomputable from Table~\ref{tab:main}); factorial rows follow the pooled five-seed protocol (Appendices F--G).}
\label{tab:ledger}
\end{table}

\begin{figure}[t]
\centering
\includegraphics[width=\columnwidth]{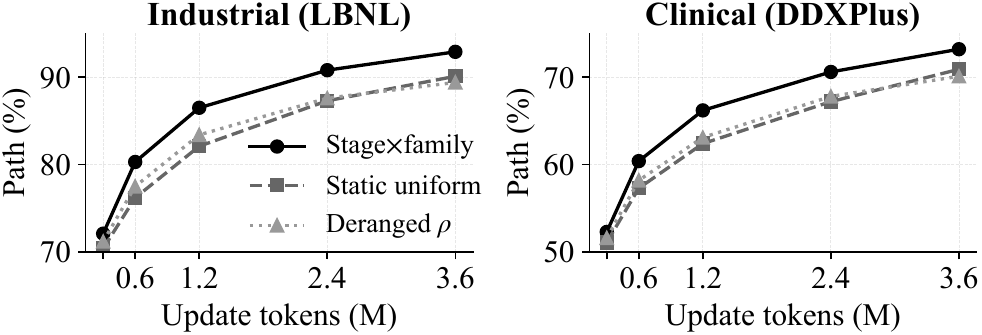}
\caption{Path versus cumulative student-update tokens (pooled over cases; Table~\ref{tab:main} reports group-macro results). ``Static uniform'' denotes the uniform-regeneration arm; markers denote frozen checkpoints; y-axis ranges differ.}
\label{fig:budget}
\end{figure}

\paragraph{Weakness-directed regeneration.}
Under the same maximum resource envelope, stage$\times$family routing reaches the prespecified stage-pass target on the development set with 64.1/58.0\% fewer generated scenarios than uniform regeneration (resource coordinates in Appendix~E). At the shared final checkpoint (equal student-update tokens), it finishes 2.8/2.3 Path points higher. The admitted corpus covers 91.4\%/88.8\% of prespecified condition--context cells and, under the frozen protocol, contains 137/84 valid combinations absent from recorded benchmark configurations. Gains concentrate where records are thinnest: the LBNL head--tail Path gap narrows from 18.9 points to 8.4, and removing any single operator family lowers Path---confounder injection the most (4.2 points), occlusion the least (1.9) (with routing variants, per-round profiles, component ablations, and $\alpha$ sensitivity, Appendices D and E). Figure~\ref{fig:budget} shows stage$\times$family routing leading at every checkpoint, while the deranged arm, ahead of uniform regeneration at mid-training, falls behind by the end.

\paragraph{Localized updating and retention.}
The episode-type$\times$reward-scope factorial (Table~\ref{tab:ledger}) crosses the rollout start (case versus repair state) with the reward scope (outcome versus criteria): repair-state episodes add 4.6/3.8 Path points, criterion rewards 3.1/2.6, and their interaction 1.9/1.5. All paired 95\% CIs exclude zero; the interaction CIs are $[0.4, 3.5]$/$[0.2, 2.9]$ (full intervals in Appendix~F). Because a repair-state episode changes both the information-bounded context and the generated continuation, the first contrast is a bundled episode effect rather than a pure start-position effect. The advantage is positive at every $g^{*}$, including 6.2/5.1 points for terminal attribution, where no repair is provided.

Continuation SFT from the same repair states, matched in examples, tokens, and updates, recovers 71\%/66\% of the gain that repair-conditioned GRPO adds over its SFT starting point: repair-state supervision itself carries most of the signal, and reinforcement learning contributes the remaining margin ($+1.35$ points on LBNL, 95\% CI $[0.4, 2.3]$) where multiple continuations are admissible. No reward-hacking signature appears in length, stuffing, or parroting checks, and equal criterion weights cost 0.6 points (Appendix~F). When repair is removed at evaluation, pass recovery at the originally failing stage is 79\%/74\%, 65\%/61\%, and 53\%/49\% for initial $g^{*}=1,2,3$ (answer-exposure and repair-content controls in Appendix~F). Preservation reduces forgetting on non-replayed strong worlds, measured per cell as the decline off its historical-best pass rate, from 9.4/7.8 to 3.4/2.9 points (component removals and the mixture sweep in Appendix~F).

\paragraph{Coupling generation and updating.}
To test whether a measured weakness must be connected to the corresponding generation operation, we keep both flywheels intact but replace the aligned map $\rho(g,v)$ with a prespecified zero-fixed-point, type-compatible derangement across cells. Measured weights and target counts stay at their original cells while each complete $\rho$ entry moves as a unit; the entries themselves, the validity standard, and the budgets are unchanged. Aligned routing improves Path over this control (Table~\ref{tab:ledger}). On DDXPlus the deranged arm still lifts Path yet leaves Acc at baseline level (78.37 vs.\ 78.44), consistent with misdirected cells still disciplining paths while no longer targeting the confusable conditions behind top-1 errors. We next cross uniform-fresh versus weakness-targeted generation with plain full-trajectory GRPO versus stage-localized updating. The interaction contrast $\Delta_{\rm int}$, the double difference across the four arms, is $+2.4$ [95\% CI 0.8, 4.1] on LBNL and $+1.9$ [0.5, 3.4] on DDXPlus for Path. The four arms share frozen cell-specific admission rules, validity standard, and budgets; the expanded definition and interaction estimates appear in Appendix~G.

\paragraph{Counterfactual generalization.}
On 100 real clinical case reports rewritten to the input schema, expert-judged Path is 64\% versus 57\% for constraint-admitted SFT, directionally consistent with the synthetic benchmarks (Appendix~B). For observation perturbations, $\mathrm{IC}_c$ requires correct answers on both members of a same-diagnosis pair; for independently constructed, training-disjoint label-changing compositions, $\mathrm{IS}_c$ additionally requires the predicted change to match the prespecified intervention. Relative to constraint-admitted SFT, DiagLoop increases $\mathrm{IC}_c$ by 8.7/7.2 points and $\mathrm{IS}_c$ by 6.3/5.4 points (conditional and excess-over-marginal analyses, with accuracy-matched and label-agnostic variants, in Appendix~G).

\section{Conclusion}

DiagLoop turns codified diagnostic mechanisms into checked training worlds without per-case expert traces; shared stage criteria govern admission, failure localization, regeneration, and reward. Across eight industrial systems and ten disease categories, the resulting 8B model improves strict path correctness by 11.6/5.5 points over the strongest conventional baseline and 3.9/2.3 over the deranged-routing control, and exceeds the evaluated proprietary references in path correctness in both domains. The method shifts rather than removes domain engineering; its benefits depend on specification coverage and checker reliability, and generalization is limited to configurations and compositions within the codified family. DDXPlus does not establish clinical readiness without a prospective clinician-in-the-loop study, and domains governed only by uncodified statistical regularities remain out of scope.

\bibliography{references}

\end{document}